\documentclass[runningheads,orivec]{llncs}

\usepackage{color,soul} 
\usepackage{makeidx}  
\usepackage{multirow}
\usepackage[T1]{fontenc}
\usepackage{graphicx}
\usepackage{subfig}
\usepackage{booktabs}
\usepackage[misc]{ifsym}
\usepackage{amsmath}

\usepackage{algorithm}
\usepackage{algpseudocode}
\usepackage{rotating}
\usepackage{pifont}
\usepackage{amssymb}
\usepackage{booktabs}
\usepackage{tabularx} 
\usepackage{enumitem}
\usepackage{xcolor}

\usepackage{xspace} 
\usepackage{listings, mdframed}

\newcommand{\name}{$ProfiLLM$\xspace}
\newcommand{\names}{$ProfiLLM^{ITSec}$\xspace}

\begin{document}

\mainmatter

\title{\name: An LLM-Based Framework for Implicit Profiling of Chatbot Users}

\titlerunning{\name: LLM-based Framework for Implicit Profiling of Chatbot Users}

\author{Shahaf David \and Yair Meidan \and Ido Hersko \and Daniel Varnovitzky  \and \\ Dudu Mimran \and Yuval Elovici \and
Asaf Shabtai}

\authorrunning{Shahaf David et al.}

\tocauthor{Shahaf David, Yair Meidan, Ido Hersko, Daniel Varnovitzky, Dudu Mimran, Yuval Elovici, Asaf Shabtai}

\institute{
Software and Information Systems Engineering Department\\Ben-Gurion University of the Negev\\Beer-Sheva, Israel\\
\email{\{sdavid, yairme, hersko, danyvarn, mimrand\}@post.bgu.ac.il,\\ \{elovici, shabtaia\}@bgu.ac.il}}

\maketitle

\begin{abstract}
Despite significant advancements in conversational AI, large language model (LLM)-powered chatbots often struggle with personalizing their responses according to individual user characteristics, such as technical expertise, learning style, and communication preferences.
This lack of personalization is particularly problematic in specialized knowledge-intense domains like IT/cybersecurity (ITSec), where user knowledge levels vary widely.
Existing approaches for chatbot personalization primarily rely on static user categories or explicit self-reported information, limiting their adaptability to an evolving perception of the user's proficiency, obtained in the course of ongoing interactions.

In this paper, we propose \name, a novel framework for implicit and dynamic user profiling through chatbot interactions.
This framework consists of a taxonomy that can be adapted for use in diverse domains and an LLM-based method for user profiling in terms of the taxonomy. 
To demonstrate \name's effectiveness, we apply it in the ITSec domain where troubleshooting interactions are used to infer chatbot users' technical proficiency.
Specifically, we developed \names, an ITSec-adapted variant of \name, and evaluated its performance on 1,760 human-like chatbot conversations from 263 synthetic users.  
Results show that \names rapidly and accurately infers ITSec profiles, reducing the gap between actual and predicted scores by up to 55--65\% after a single prompt, followed by minor fluctuations and further refinement. 
In addition to evaluating our new implicit and dynamic profiling framework, we also propose an LLM-based persona simulation methodology, a structured taxonomy for ITSec proficiency, our codebase, and a dataset of chatbot interactions to support future research.

\keywords {User profiling, large language models (LLMs), artificial intelligence (AI), chatbots, cybersecurity, taxonomy}
\end{abstract}

\section{Introduction}\label{sec:Introduction}

Large language models (LLMs) have transformed the capabilities of conversational AI, advancing chatbots and overcoming their previous limitations to facilitate more nuanced and contextually aware interactions~\cite{panagiotidis2024llm}.
In contrast to traditional rule-based chatbots with constrained response patterns, LLM-based conversational agents demonstrate enhanced context comprehension and linguistic adaptability.
The rapid adoption of this technology across diverse industries, with implementations in customer service, healthcare, user profiling domains, and more~\cite{meduri10adaptive,santosh2024}, is a reflection of its transformative potential, and market analysis projects exponential growth in the coming years~\cite{gartner2023b}.

LLM-powered chatbots excel in natural language processing but often fail to tailor responses to users' technical proficiency, learning styles, or communication preferences~\cite{zhang2024chatbots}.  
This lack of personalization is especially problematic in specialized domains with varying user knowledge levels.  
In professional and educational settings, misalignment between user expertise and chatbot responses can lead to frustration and reduced engagement~\cite{zhang2024chatbots}. 
For example, technical support chatbots may overwhelm novices with complex jargon or oversimplify explanations for experts, impacting these chatbots' ability to achieve their intended purposes.

Recent work on conversational AI personalization has explored dialogue style adaptation and expertise-based response generation~\cite{cao2023diaggpt,hudevcek2023llms}.  
However, most approaches rely on \emph{explicit} user questionnaires or \emph{static} predefined profile categories, lacking \emph{implicit} (manual input-free) and \emph{dynamic} (adaptive) profiling.
To address this gap, we introduce \emph{\name}, a framework for continuous user profiling through chatbot interactions.  
\name features a domain-adaptable taxonomy and an LLM-based inference method that profiles users implicitly, without questionnaires or direct assessments.  

For evaluation, we adapt \name for IT/cybersecurity (ITSec), where user proficiency shapes troubleshooting conversations.  

Specifically, we built an ITSec-oriented chatbot with a \names module.  
We also designed a technical proficiency questionnaire, completed by 63 human responders, to establish ITSec profile archetypes.  
From these archetypes, we generated 352 synthetic users via random sampling with noise.  
After rigorous data refinement, we evaluated \names using 1,315 high-quality chatbot conversations on ITSec troubleshooting.  
Results show that with an optimized configuration (Sec.~\ref{subsec:Hyperparameter_Optimization_LLM_beta_and_W}), \names achieves a rapid accuracy boost, improving profiling by 55–65\% after just one prompt, with further gains thereafter.

The contributions of our work can be summarized as follows:
\begin{enumerate}[nosep, leftmargin=*]
    \item We introduced a novel LLM-based framework for dynamic, fully implicit user profiling based solely on chatbot interactions.  
    \item We proposed a structured taxonomy for ITSec proficiency modeling and developed a corresponding questionnaire for user assessment.  
    \item We presented an LLM-based approach for persona generation, persona-driven chatbot interaction simulation, and data quality assurance.  
    \item We conducted a rigorous empirical evaluation of all components and share our findings.  
    \item We compiled a dataset of ITSec troubleshooting conversations, labeled with users' ITSec profiles.  
    To support further research, we publicly release this dataset and related code\footnote{Repository URL to be shared upon request}.  
\end{enumerate}

\section{Proposed Method}\label{sec:Proposed_Method}

\name comprises two key components, which are detailed  next: (1) the taxonomy generation component, in which a domain-specific taxonomy is generated, and (2) the profile inference component, in which a user profile (in terms of the taxonomy) is inferred.

\subsection{Domain-Adapted Taxonomy Generation}\label{subsec:Domain_Adapted_axonomy_Generation}

To ensure consistency and preciseness across all user interactions, \name takes a structured approach to user profiling.
In practice, the user's profile $P$ (i.e., the user's knowledge or proficiency) is hierarchically broken down into $n$ subdomains $p_1, p_2, ..., p_n \in P$. 
These $n$ low-level subdomains are the profile elements directly inferred (scored) by \name, while the higher-level domains in the taxonomy enable logical categorization.
A domain-specific list of profile subdomains should aim at maximum comprehensiveness so that as many chatbot interactions as possible would be both assigned to at least one relevant subdomain and directly scored.
At the same time, the subdomain list should not be over-complicated, making the taxonomy sparse and impractical. 

We were unable to identify any taxonomy that is relevant to the domain of ITSec. 
To address this gap, we created a novel taxonomy based on a thorough review of various relevant sources, including syllabuses of IT support courses offered by IBM, Google, and Dell on Coursera~\cite{coursera
}, governmental resources, research papers, technological forums, surveys on prevalent ITSec complaints, online tutorials and blogs~\cite{electric2025}. 
The ITSec taxonomy we created, consisting of 23 subdomains grouped by five domains, is presented in Table~\ref{tab:domains_subdomains}.
\begin{table}[ht]
    \centering
    \caption{Our proposed taxonomy for ITSec proficiency profiling.}
    \label{tab:domains_subdomains}
    \scriptsize 
    \resizebox{\textwidth}{!}{
    \begin{tabular}{|c|l|l|}
        \hline
        \textbf{Domain} & \textbf{Subdomain} & \textbf{Example of related user complaint} \\ 
        \hline
        \multirow{4}{*}{\shortstack[c]{Hardware\\(HW)}} & Hardware - General & "My laptop won't turn on; what should I do?" \\ 
        & Peripherals & "My external keyboard is not working; how can I fix it?" \\ 
        & Storage & "How much free space do I have on my hard drive?" \\ 
        & RAM and Memory & "How can I check if I need to upgrade my RAM?" \\ 
        \hline
        \multirow{5}{*}{\shortstack[c]{Networking\\(NT)}} & Networking - General & "Why is my Internet connection so slow?" \\ 
        & Cloud Networking & "How can I access my cloud storage from another device?" \\ 
        & Protocols & "What is the difference between TCP and UDP?" \\ 
        & Configuration & "How do I change my router's Wi-Fi password?" \\ 
        & Security & "How can I protect my home network from hackers?" \\ 
        \hline
        \multirow{6}{*}{\shortstack[c]{Cybersecurity\\(CS)}} & Cybersecurity - General & "What are the best practices for online security?" \\ 
        & Data Leakage & "How do I prevent my personal data from being leaked?" \\ 
        & Privacy & "How can I browse the Internet without being tracked?" \\ 
        & Malware & "I think my computer has a virus; what should I do?" \\ 
        & Encryption & "How can I encrypt my text file for security?" \\ 
        & Authentication & "How can I enable two-factor authentication for my mail?" \\ 
        \hline
        \multirow{4}{*}{\shortstack[c]{Software\\(SW)}} & Software - General & "How do I install a new application on my computer?" \\ 
        & App Management & "Why is my app crashing every time I open it?" \\ 
        & Programming & "How do I fix a syntax error in Python?" \\ 
        & Web Browsers & "How do I clear my browsing history in Chrome?" \\ 
        \hline
        \multirow{4}{*}{\shortstack[c]{Operating\\Systems\\(OS)}} & Operating Systems - General & "What's the difference between Windows and Linux?" \\ 
        & Drivers & "How do I update my graphics card driver?" \\ 
        & File Management & "How do I restore a deleted file from the recycle bin?" \\ 
        & Settings and Configurations & "How do I change the default language in Windows?" \\ 
        \hline
    \end{tabular}
    }
\end{table}

\subsection{User Profile Inference in Terms of the Taxonomy}\label{subsec:Profile_Inference}

\begin{figure}[ht]
    \centering
    \includegraphics[width=0.95\textwidth]{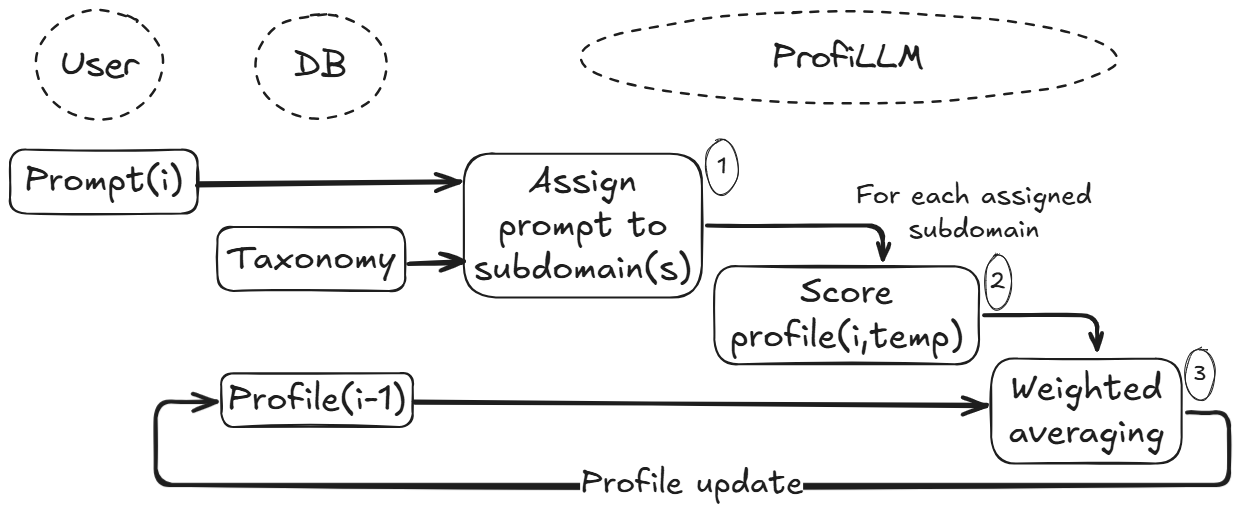}
    \caption{User Profile Inference in Terms of the Taxonomy}
    \label{fig:profile_inference_diagram}
\end{figure}

This component of \name infers the user profile in various taxonomy subdomains based on their interactions with the conversational interface.
As shown in Fig.~\ref{fig:profile_inference_diagram}, for every prompt submitted by the user, (1) the prompt is assigned to one or more of the taxonomy subdomains, (2) the prompt and its context are separately given a score for each assigned subdomain, and (3) these new scores are weighted against the existing related scores to update the profile scores.

\subsubsection{Prompt Assignment to Taxonomy Subdomains.}\label{subsubsec:Prompt_Assignment_to_Taxonomy_Subdomains}

Each conversational iteration (interaction) $i$ between a user $u$ and the chatbot begins with $prompt_i^u$.
Given $prompt_i^u$ and its context window $W_i^u$ (the preceding \{user prompt, chatbot response\} pairs), \name uses crafted system prompts to identify relevant subdomains in the taxonomy.
This contextual and focused analysis ensures that the profile assessment is tied to the particular areas discussed by the user.

\subsubsection{Subdomain-Specific Profile Scoring.}\label{subsubsec:Subdomain_Specific_Profile_Scoring}

For each assigned subdomain (profile element) $p^j$, \name assesses the chatbot user's knowledge level on a discrete five-point scale, where higher scores indicate greater expertise. For instance, a prompt such as \textit{"How can I decrypt .locky files encrypted by ransomware?"} would be categorized under "Cybersecurity/Malware" and temporarily scored as $p^{CS/Malware, temp}_i=5$. Conversely, \textit{"My PC wants money to free my locked files"} would assigned to the same subdomain but scored much lower, e.g., $p^{CS/Malware, temp}_i=2$.
This scoring mechanism evaluates (1) concept complexity, (2) terminology appropriateness, (3) depth of understanding in $prompt_i^u$'s formulation, and (4) contextual relevance to prior interactions.

\subsubsection{Score Aggregation via Weighted Averaging.}\label{subsubsec:Score_Aggregation_and_Confidence_Weighting}

In each iteration $i$, for every assigned subdomain $j$, \name updates $p^j_i$ via weighted averaging (Eq.~\ref{eq:score_update}):

\begin{equation}\label{eq:score_update}
    p^j_i = \alpha^j_i \cdot p^{j, temp}_i + (1 - \alpha^j_i) \cdot p^j_{i-1}
\end{equation}

Here, the temporary score $p^{j, temp}_i$ is weighted by $\alpha_i^j$, while the previous score $p^j_{i-1}$ is weighted by $1-\alpha_i^j$. 
Since $\alpha_i^j$ is dynamic, early interactions prioritize new input ($\alpha_i^j$ is higher), ensuring rapid adaptation when the existing profile score $p^j_{i-1}$ is less reliable.  
As confidence in $p^j_{i-1}$ increases, $\alpha^j_i$ gradually decreases, stabilizing the profile score.
The confidence weight $\alpha^j_i$ follows an inverse time-decay function (Eq.~\ref{eq:alpha_update}), adjusting based on the number of interactions $i$:

\begin{equation}\label{eq:alpha_update}
\alpha^j_i = \frac{\alpha_0}{1 + \beta \cdot i}
\end{equation}

In our experiments, we optimize (1) the decay rate $\beta$, which controls how quickly $\alpha_i^j$ stabilizes, and (2) the context window size $|W|$, i.e., the maximum number of recent \{user prompt, chatbot response\} pairs used for scoring.

\section{Evaluation Method}\label{sec:Evaluation_Method}

We empirically evaluated the generic \name framework by (1) implementing an ITSec troubleshooting chatbot and a \names instance, (2) curating dozens of ground-truth human profiles, (3) generating hundreds of synthetic users, seeded using cluster centroids of the human profiles, and (4) analyzing thousands of chatbot interactions held by the synthetic users.
We curated (rather than assumed) ground-truth human profiles to make the evaluation realistic.
Inspired by previous research~\cite{tamoyan2024llm}, we leveraged these profiles (persona archetypes) by generating numerous comparable synthetic users, which automatically held a large number of human-like ITSec-related conversations with the chatbot.

\subsection{\names Implementation}\label{subsec:ProfiLLM_Implementation}

Our experiments were conducted on Azure, providing access to various pretrained LLMs (Sec.~\ref{subsec:Hyperparameter_Optimization_LLM_beta_and_W}).  
We used OpenAI's GPT-4o as the backend for our ITSec troubleshooting chatbot, developed within this setup.  
Both the chatbot and \names were implemented using the LangChain framework.

\subsection{Human Ground-Truth Labeling}\label{subsec:Human_Ground_Truth_Labeling}

To curate ground-truth human profiles, we developed an ITSec proficiency questionnaire.
For each of the 23 subdomains in the taxonomy (Table~\ref{tab:domains_subdomains}), this online questionnaire\footnote{Questionnaire URL to be shared upon request.} assesses users' proficiency level using three types of questions:
\begin{enumerate}[nosep, leftmargin=*]
    \item \emph{Self-assessment questions} (1--5 scale): Evaluating the user's \emph{perceived} proficiency, e.g., \textit{"How would you rate your knowledge on network configuration?"}
    \item \emph{Conceptual questions} (1--5 scale): Measuring the user's understanding of technical concepts, e.g., \textit{"How well do you understand DHCP and DNS services?"}
    \item \emph{Practical questions} (Binary, 1 or 5): Assessing the ability to apply knowledge in real scenarios, e.g., \textit{"Have you configured static IP addresses on devices?"}
\end{enumerate}

Our questionnaire's 1--5 scale is based on the EU's DigComp~\cite{vuorikari2022digcomp}.  
It retains the original proficiency levels -- \textit{foundation} (2), \textit{intermediate} (3), \textit{advanced} (4), and \textit{highly specialized} (5) -- with an additional lower level, \textit{no knowledge} (1).  
A subdomain's final score is the average of the three corresponding questionnaire responses.  
The user's ITSec profile is then represented as a numeric vector, where each 1--5 entry indicates their proficiency in a specific subdomain.

\subsection{Generation of Human-Like Synthetic Users}\label{subsec:Synthetic_User_Generation}

After curating a variety of human user profiles, we performed cluster analysis and used the resulting centroids as ITSec persona archetypes.
The first step in generating a synthetic user involved randomly selecting one of these centroid vectors and adding random noise (from a uniform distribution) to each subdomain score.  
Preliminary experimentation showed that directly using the numeric profile vector often produces synthetic conversations that lack human-like quality, making them unrepresentative.  
To address this, we developed a prompt that converts a numeric profile vector into a textual persona description.  
For low-proficiency profiles, this prompt includes the following text:
\textit{"... Create a detailed, realistic human persona... based on the technical proficiency profile... Consider personas like... traditional craftspeople who work entirely manually, ... people who prefer face-to-face interactions and paper-based systems... reflect their technical proficiency level through a realistic and engaging narrative...(400-500 words):
[Name] is a [age]-year-old [traditional profession] who excels at [hands-on skill/community role]... rely entirely on [traditional methods/tools] to [accomplish goals]... When faced with modern technology, they [coping strategy that emphasizes reliance on family/community]..."}.
Like previous research~\cite{abbasiantaeb2024let}, a detailed persona description was created to enhance LLM-generated conversations, however we add the challenge of having a long numeric profile vector as a starting point.

\subsection{Conversation Data Collection and Refinement}\label{subsec:Conversation_Data_Collection_and_Refinement}

Each generated synthetic user was tasked with solving ITSec scenarios exclusively through interactions with our troubleshooting chatbot:  
\begin{enumerate}[nosep, leftmargin=*]
    \item HW -- Flickering monitor, jerky mouse movements, unresponsive keyboard  
    \item NT -- Files on shared drives occasionally corrupt  
    \item CB -- Suspicious calendar invites from colleagues who deny sending them  
    \item SW -- Programs freeze when switching windows or tabs  
    \item OS -- System notifications appear delayed or out of order  
\end{enumerate}

The (synthetic) users were instructed to maintain consistent behavior and phrase their prompts strictly in accordance with their assigned persona.
Upon completion, another LLM valuated the conversations based on two criteria: the alignment between the user's profile and their behavior/phrasing and the naturalness of the conversation flow, i.e., its resemblance to human interactions. 
Conversations not meeting either criterion were excluded from the dataset.

\subsection{Performance Metrics}\label{subsec:Performance_Metrics}

For a subdomain, the absolute error (AE), representing the difference between the actual and inferred profile scores, serves as the simplest performance metric.
For a domain, or an entire profile, we measure \names's performance using the mean absolute error (MAE), assuming equal subdomain weights.
An ideal model's MAE would quickly converge to zero, i.e., close the gap between the actual and predicted profile scores using only a small sequence of user prompts.

\section{Results}\label{sec:Results}

In this section, we present a data overview and hyperparameter tuning, followed by progressive evaluation of various aspects using the optimal configuration.

\subsection{Experimental Data Collected}\label{subsec:Experimental_Data_Collected}

\begin{figure}[ht]
    \centering
    \subfloat[Cluster centroids (persona archetypes)]{
        \includegraphics[width=0.50\textwidth, trim={0 35 10 10},clip]{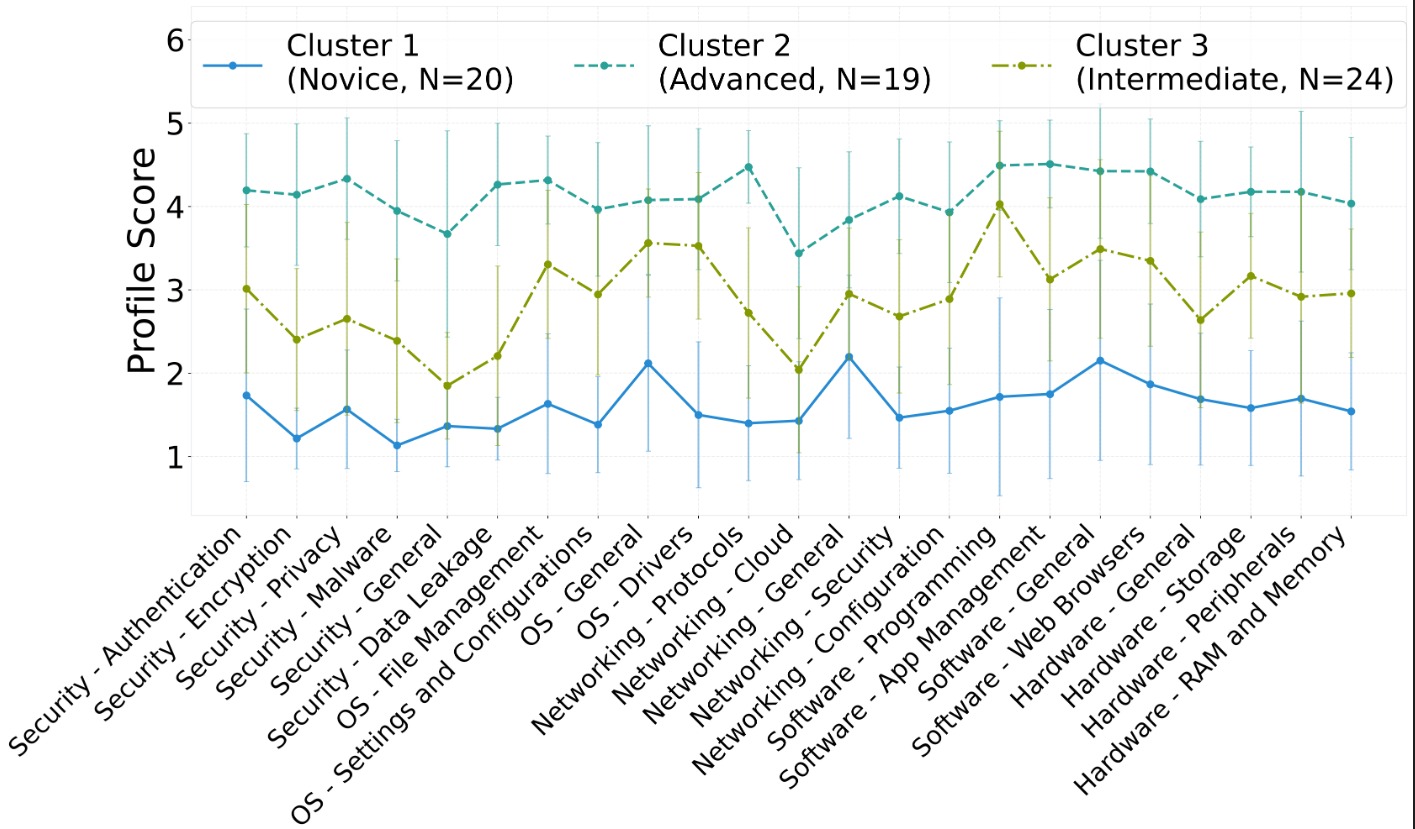}
        \label{fig:centroids}
    }
    \hfill
    \subfloat[Conversation quality distribution]{
        \includegraphics[width=0.46\textwidth]{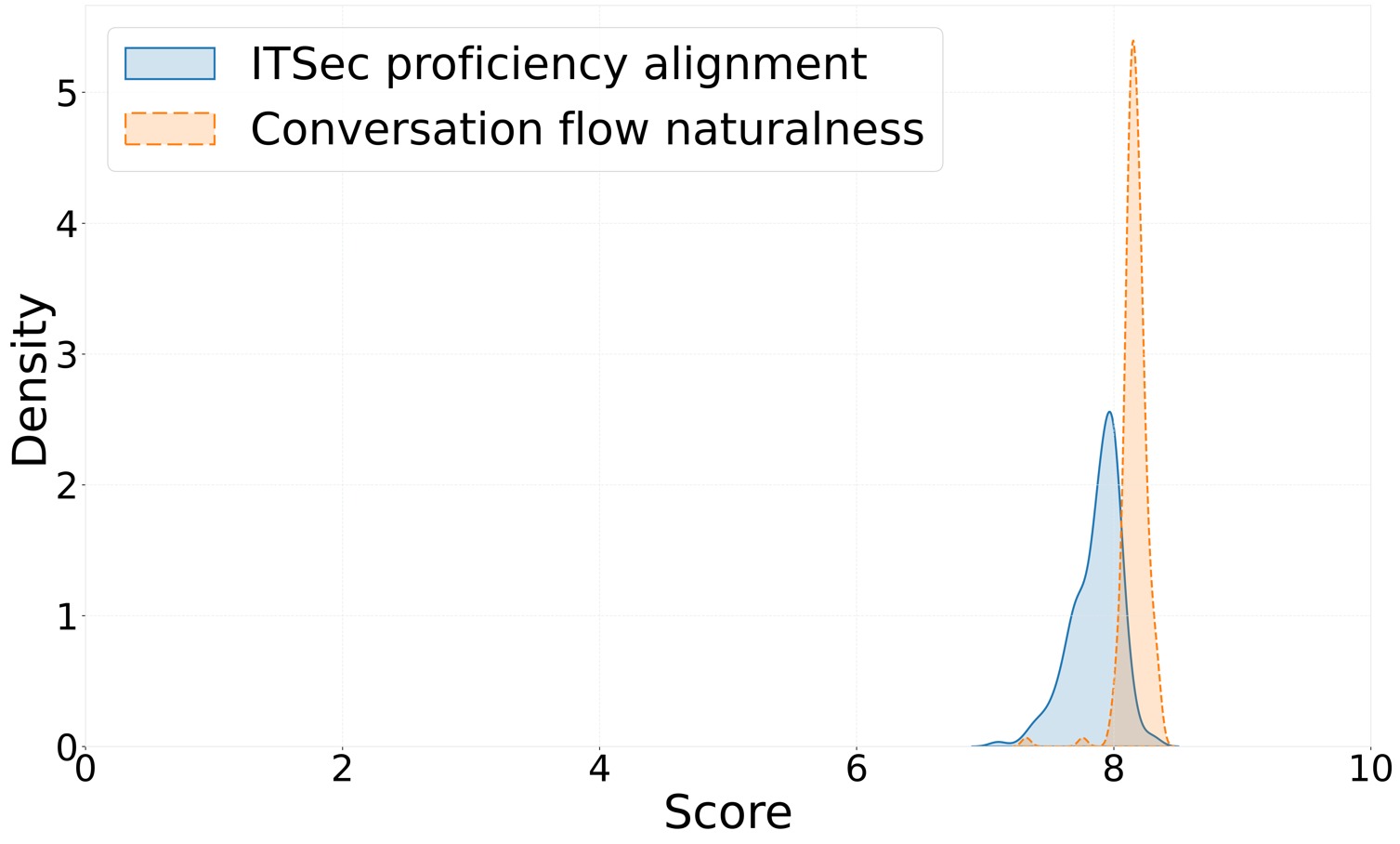}
        \label{fig:naturalness}
    }
    \caption{Experimental data collected: cluster centroids and conversation quality}
    \label{fig:combined}
\end{figure}

Our ITSec questionnaire (Sec.~\ref{subsec:Human_Ground_Truth_Labeling}) was completed by 63 participants from diverse academic and vocational backgrounds.  
$k$-Means clustering ($k=3$) yielded the most consistent user groupings, supported by the Elbow Method and Silhouette Score analysis.  
The resulting centroids (Fig.~\ref{fig:centroids}) define three stable ITSec persona archetypes: \textit{advanced} (high scores across most subdomains), \textit{intermediate}, and \textit{novice}.  
Using these centroids, 352 synthetic users were generated (Sec.~\ref{subsec:Synthetic_User_Generation}), engaging in 1760 ITSec-related conversations, assessed for quality using Claude.  
Most were rated highly for profile alignment and conversational flow (Fig.~\ref{fig:naturalness}).  
Applying an 8.5 quality threshold, the final dataset comprised 1315 conversations from 263 users.  
The associated data processing overhead is detailed in Sec.~\ref{subsec:Overhead}.

\subsection{Hyperparameter Optimization: $LLM$, $\beta$ and $|W|$}\label{subsec:Hyperparameter_Optimization_LLM_beta_and_W}

Table~\ref{tab:mae-combinations} presents the results of a grid search conducted using a sample of conversations from 30 randomly selected users.
Our goal was to identify the optimal configuration -- the combination of $LLM$, $\beta$, and $|W|$ -- that minimizes the MAE.
Note that the MAE at iteration $i$ ($MAE@i$) is not calculated chronologically based on the \emph{overall} prompt sequence.
Instead, since a subdomain's score is updated \emph{only} when a prompt is specifically assigned to it, $MAE@i$ corresponds to the $i^{th}$ prompt \emph{assigned within each subdomain}. 
For example, $MAE@2$ considers only the AEs computed at the second prompt assigned to each of the 23 profile subdomains.
Thus, $MAE@2$ answers the question: \textit{For any given subdomain, what is the expected AE after exactly two prompts have been assigned to it?}

\begin{table}[!ht]
\centering
 \caption{Grid search over $LLM$, $\beta$ and $|W|$ to find the optimal combination that minimizes the MAE. Results are displayed as Mean$\pm$St.Dev.}
\label{tab:mae-combinations}
\scriptsize
\resizebox{\columnwidth}{!}{%
\begin{tabular}{lcc|cccccc}
\hline
\textbf{$LLM$} & \textbf{$\beta$} & \textbf{$|W|$} & \textbf{$MAE@0$} & \textbf{$MAE@1$} & \textbf{$MAE@2$} & \textbf{$MAE@3$} & \textbf{$MAE@4$} & \textbf{$MAE@5$} \\
\hline
\multirow{9}{*}{GPT-4o mini} 
& 0.1 & 0 & 1.14$\pm$0.38 & 0.62$\pm$0.52 & 0.83$\pm$0.55 & 0.64$\pm$0.50 & 0.71$\pm$0.51 & 0.63$\pm$0.48 \\
& 0.1 & 1 & 1.14$\pm$0.38 & 0.63$\pm$0.53 & 0.86$\pm$0.58 & 0.63$\pm$0.50 & 0.72$\pm$0.52 & 0.61$\pm$0.46 \\
& 0.1 & 999 & 1.14$\pm$0.38 & 0.63$\pm$0.53 & 0.81$\pm$0.58 & 0.62$\pm$0.50 & 0.68$\pm$0.52 & 0.63$\pm$0.50 \\
& 0.2 & 0 & 1.14$\pm$0.38 & 0.64$\pm$0.55 & 0.78$\pm$0.57 & 0.66$\pm$0.52 & 0.77$\pm$0.51 & 0.59$\pm$0.46 \\
& 0.2 & 1 & 1.14$\pm$0.38 & 0.63$\pm$0.54 & 0.82$\pm$0.57 & 0.68$\pm$0.52 & 0.73$\pm$0.52 & 0.64$\pm$0.49 \\
& 0.2 & 999 & 1.14$\pm$0.38 & 0.62$\pm$0.53 & 0.80$\pm$0.58 & 0.68$\pm$0.54 & 0.72$\pm$0.51 & 0.62$\pm$0.48 \\
& 0.3 & 0 & 1.14$\pm$0.38 & 0.64$\pm$0.54 & 0.79$\pm$0.57 & 0.65$\pm$0.52 & 0.67$\pm$0.50 & 0.63$\pm$0.47 \\
& 0.3 & 1 & 1.14$\pm$0.38 & 0.62$\pm$0.54 & 0.82$\pm$0.56 & 0.61$\pm$0.50 & 0.77$\pm$0.51 & 0.65$\pm$0.48 \\
& 0.3 & 999 & 1.14$\pm$0.38 & 0.63$\pm$0.53 & 0.81$\pm$0.56 & 0.63$\pm$0.50 & 0.71$\pm$0.51 & 0.60$\pm$0.48 \\
\hline

\multirow{9}{*}{Claude 3.5 sonnet} 
& 0.1 & 0 & 1.14$\pm$0.38 & 0.69$\pm$0.58 & 0.88$\pm$0.55 & 0.75$\pm$0.54 & 0.85$\pm$0.54 & 0.72$\pm$0.48 \\
& 0.1 & 1 & 1.14$\pm$0.38 & 0.71$\pm$0.59 & 0.85$\pm$0.53 & 0.67$\pm$0.50 & 0.77$\pm$0.53 & 0.69$\pm$0.48 \\
& 0.1 & 999 & 1.14$\pm$0.38 & 0.73$\pm$0.60 & 0.86$\pm$0.58 & 0.73$\pm$0.55 & 0.82$\pm$0.54 & 0.70$\pm$0.51 \\
& 0.2 & 0 & 1.14$\pm$0.38 & 0.69$\pm$0.58 & 0.90$\pm$0.58 & 0.75$\pm$0.57 & 0.86$\pm$0.54 & 0.70$\pm$0.52 \\
& 0.2 & 1 & 1.14$\pm$0.38 & 0.73$\pm$0.59 & 0.93$\pm$0.57 & 0.68$\pm$0.50 & 0.83$\pm$0.53 & 0.68$\pm$0.51 \\
& 0.2 & 999 & 1.14$\pm$0.38 & 0.70$\pm$0.59 & 0.86$\pm$0.57 & 0.73$\pm$0.54 & 0.83$\pm$0.54 & 0.71$\pm$0.50 \\
& 0.3 & 0 & 1.14$\pm$0.38 & 0.70$\pm$0.59 & 0.87$\pm$0.56 & 0.72$\pm$0.51 & 0.78$\pm$0.51 & 0.69$\pm$0.49 \\
& 0.3 & 1 & 1.14$\pm$0.38 & 0.68$\pm$0.58 & 0.94$\pm$0.54 & 0.60$\pm$0.49 & 0.84$\pm$0.49 & 0.62$\pm$0.46 \\
& 0.3 & 999 & 1.14$\pm$0.38 & 0.71$\pm$0.60 & 0.86$\pm$0.57 & 0.77$\pm$0.58 & 0.81$\pm$0.58 & 0.74$\pm$0.54 \\
\hline
\multirow{9}{*}{Llama 3.3 70b} 
& 0.1 & 0 & 1.14$\pm$0.38 & 0.59$\pm$0.44 & 0.82$\pm$0.49 & 0.61$\pm$0.44 & 0.76$\pm$0.45 & 0.64$\pm$0.45 \\
& 0.1 & 1 & 1.14$\pm$0.38 & 0.56$\pm$0.42 & 0.72$\pm$0.48 & 0.64$\pm$0.48 & 0.71$\pm$0.48 & 0.63$\pm$0.46 \\
& 0.1 & 999 & 1.14$\pm$0.38 & 0.57$\pm$0.42 & 0.79$\pm$0.49 & 0.66$\pm$0.46 & 0.79$\pm$0.45 & 0.65$\pm$0.47 \\
& 0.2 & 0 & 1.14$\pm$0.38 & 0.52$\pm$0.41 & 0.77$\pm$0.52 & 0.58$\pm$0.45 & 0.66$\pm$0.43 & 0.62$\pm$0.45 \\
& 0.2 & 1 & 1.14$\pm$0.38 & 0.57$\pm$0.41 & 0.71$\pm$0.49 & 0.67$\pm$0.43 & 0.74$\pm$0.45 & 0.64$\pm$0.42 \\
& 0.2 & 999 & 1.14$\pm$0.38 & 0.55$\pm$0.41 & 0.76$\pm$0.46 & 0.60$\pm$0.43 & 0.73$\pm$0.48 & 0.57$\pm$0.42 \\
& 0.3 & 0 & 1.14$\pm$0.38 & 0.51$\pm$0.41 & 0.74$\pm$0.53 & 0.66$\pm$0.48 & 0.73$\pm$0.45 & 0.59$\pm$0.41 \\
& 0.3 & 1 & 1.14$\pm$0.38 & 0.56$\pm$0.38 & 0.76$\pm$0.48 & 0.63$\pm$0.45 & 0.74$\pm$0.44 & 0.61$\pm$0.44 \\
& 0.3 & 999 & 1.14$\pm$0.38 & 0.56$\pm$0.40 & 0.79$\pm$0.49 & 0.60$\pm$0.42 & 0.80$\pm$0.45 & 0.60$\pm$0.42 \\
\hline
\multirow{9}{*}{GPT-4o} 
& 0.1 & 0 & 1.14$\pm$0.38 & 0.51$\pm$0.35 & \textbf{0.64$\pm$0.48} & 0.57$\pm$0.45 & 0.61$\pm$0.48 & 0.60$\pm$0.48 \\
& 0.1 & 1 & 1.14$\pm$0.38 & 0.54$\pm$0.41 & 0.71$\pm$0.48 & \textbf{0.53$\pm$0.43} & 0.65$\pm$0.49 & 0.53$\pm$0.44 \\
& 0.1 & 999 & 1.14$\pm$0.38 & \textbf{0.48$\pm$0.38} & 0.71$\pm$0.51 & 0.57$\pm$0.46 & 0.62$\pm$0.47 & 0.55$\pm$0.43 \\
& 0.2 & 0 & 1.14$\pm$0.38 & 0.54$\pm$0.36 & 0.67$\pm$0.47 & 0.62$\pm$0.46 & 0.70$\pm$0.48 & 0.56$\pm$0.42 \\
& 0.2 & 1 & 1.14$\pm$0.38 & 0.53$\pm$0.38 & 0.78$\pm$0.45 & 0.57$\pm$0.42 & 0.67$\pm$0.47 & \textbf{0.50$\pm$0.40} \\
& 0.2 & 999 & 1.14$\pm$0.38 & 0.51$\pm$0.36 & 0.71$\pm$0.51 & 0.62$\pm$0.43 & 0.67$\pm$0.48 & 0.61$\pm$0.44 \\
& 0.3 & 0 & 1.14$\pm$0.38 & 0.51$\pm$0.34 & 0.71$\pm$0.52 & 0.56$\pm$0.43 & 0.72$\pm$0.47 & 0.63$\pm$0.43 \\
& 0.3 & 1 & 1.14$\pm$0.38 & 0.49$\pm$0.38 & 0.72$\pm$0.47 & 0.54$\pm$0.43 & \textbf{0.59$\pm$0.45} & 0.55$\pm$0.41 \\
& 0.3 & 999 & 1.14$\pm$0.38 & 0.54$\pm$0.38 & 0.69$\pm$0.50 & 0.60$\pm$0.48 & 0.62$\pm$0.45 & 0.59$\pm$0.46 \\
\hline
\multicolumn{9}{l}{\textit{Note: Lower values indicate better performance (smaller error).}} \\
\end{tabular}
}
\end{table}

\begin{table}[t]
\centering
\caption{Distribution of subdomain-specific user prompt sequence lengths}
\label{tab:sequence_length}
\resizebox{\columnwidth}{!}{%
\begin{tabular}{lrrrrrrrrrrrrr}
\toprule
& \textbf{Mean} & \textbf{St.Dev.} & \textbf{Min} & \textbf{10\%} & \textbf{20\%} & \textbf{30\%} & \textbf{40\%} & \textbf{50\%} & \textbf{60\%} & \textbf{70\%} & \textbf{80\%} & \textbf{90\%} & \textbf{Max} \\
\midrule
\textbf{Sequence length} & 5.39 & 4.93 & 1.00 & 1.00 & 1.00 & 1.00 & 1.00 & 4.00 & 6.00 & 8.00 & 10.00 & 13.00 & 19.00 \\
\bottomrule
\end{tabular}%
}
\end{table}

Using $\alpha_0=0.8$, Table~\ref{tab:mae-combinations} shows that within the first five iterations, GPT-4o consistently outperforms all other evaluated LLMs, predominantly with a context window of length 1.
In iterations 1-3, $\beta=0.1$ also consistently leads to the lowest MAE.
Although in iterations 4-5 other $\beta$ values perform better, we decided to set the optimal configuration \names$^*$ as $LLM^*=GPT-4o$, $|W|^*=1$ and $\beta^*=0.1$.
The reason for doing so is evident in Table~\ref{tab:sequence_length}, which shows that half of the subdomain-specific user prompt sequences are 4 or less in length.
That is, since interactions on specific subdomain are relatively short, the practical optimality of the hyperparameter configuration ($LLM^*$, $|W|^*$ and $\beta^*$) depends on the decrease in MAE during the first few interactions.

\begin{figure}[!b]
    \centering
    \includegraphics[width=1.0\textwidth]{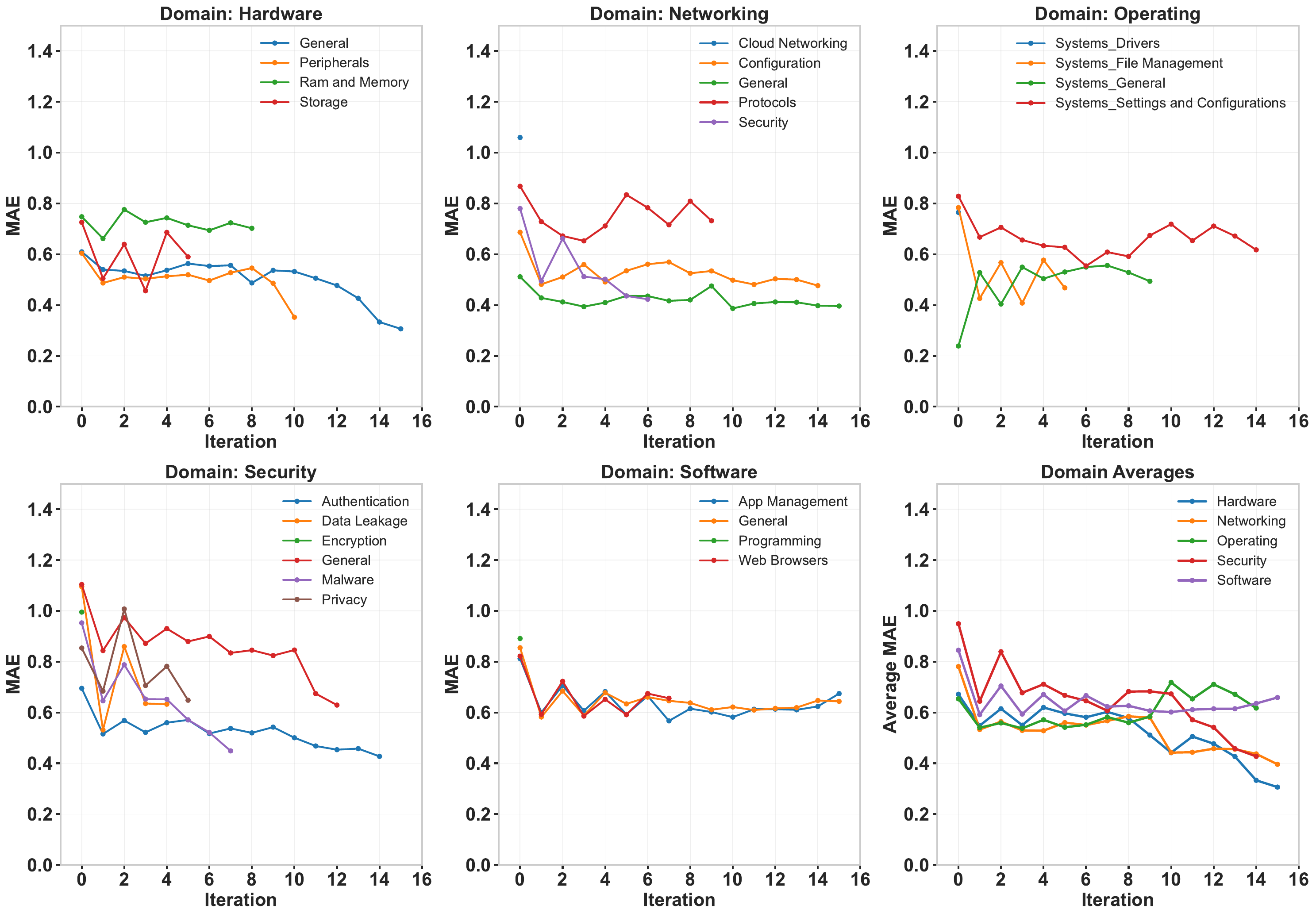}
    \caption{\names$^*$'s performance across profile domains and subdomains}    \label{fig:five_plus_one_MAE_domains_and_subdomains}
\end{figure}

\subsection{Performance across Domains and Subdomains}\label{subsec:Performance_across_Domains_and_Subdomains}

After optimizing the hyperparameters using a random sample of users and chatbot interactions, we applied \names$^*$ to the entire refined experimental dataset (Sec.~\ref{subsec:Experimental_Data_Collected}), evaluating its performance across all domains and subdomains. As illustrated in Fig.~\ref{fig:five_plus_one_MAE_domains_and_subdomains}, the MAE performed as anticipated in most cases:
\begin{itemize}[nosep, leftmargin=*]
    \item In iteration 0, where without any prior user information we assume average ITSec proficiency ($p=3, \quad \forall p \in P$), typically $MAE<1.0$, meaning that the actual proficiency scores range mainly between 2.0 and 4.0.
    \item In almost all cases, $MAE@1$ shows a marked improvement in profile inference, with  approximately 30\% reduction compared to $MAE@0$, on average.
    \item During the first few iterations, the MAE fluctuates, probably due to the relatively high $\alpha_i$ of the temporary profile scores $p_i^{j,temp}$ when $i$  is low (Eq.~\ref{eq:alpha_update}).
    \item Later on, the gradual decrease in $\alpha_i$ is reflected via diminishing bumpiness in MAE, which eventually converges to $[0.3, 0.65]$, on average per domain.
    Practically, a profiling error of this range is negligible for the intended response adaptation (e.g., adjusted jargon and complexity).  
    For instance, a prompt instructing a user to check their free disk space would be phrased similarly for users with $p^{HW/Storage}$ of 3.3 or 3.8.
\end{itemize}

Although the previously described pattern -- a marked decrease in MAE, followed by fluctuations and gradual convergence -- is generally evident, performance varied across subdomains, with some performing better and others worse.
To investigate these differences, we (1) evaluated \names$^*$'s performance across the three persona archetypes, (2) examined the effect of prompt length, (3) conducted an ablation study, (4) experimented with human users, and (5) qualitatively analyzed a few conversations, as elaborated next. 

\subsection{Performance across Persona Archetypes}\label{subsec:Performance_across_Persona_Archetypes}

\begin{figure}[ht]
    \centering
    \subfloat[MAE across persona archetypes]{
        \includegraphics[width=0.48\textwidth, trim={0 7 0 11},clip]{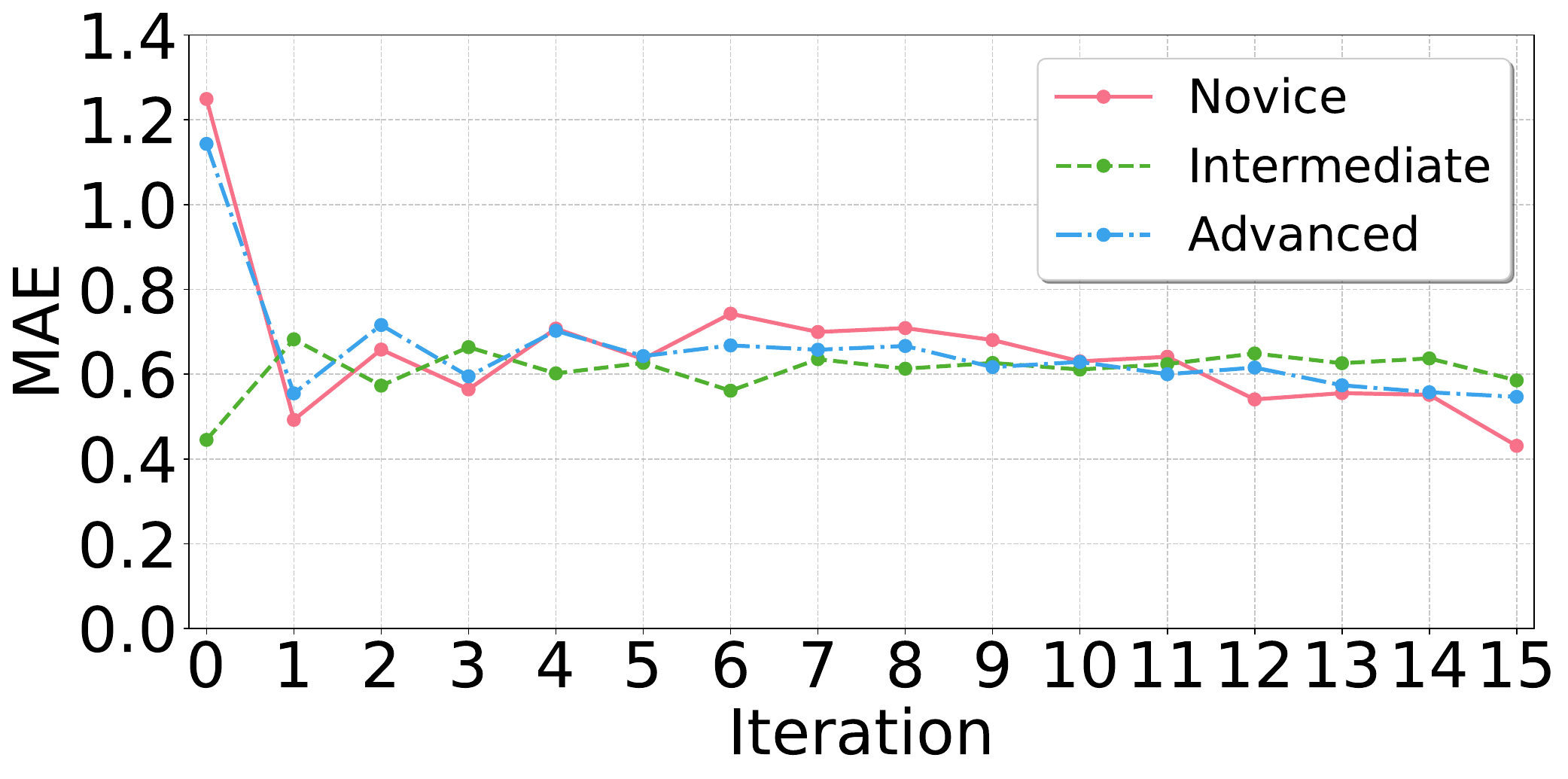}
        \label{fig:MAE_across_persona_archetypes}
    }
    \hfill
    \subfloat[Effect of prompt length]{
        \includegraphics[width=0.46\textwidth]{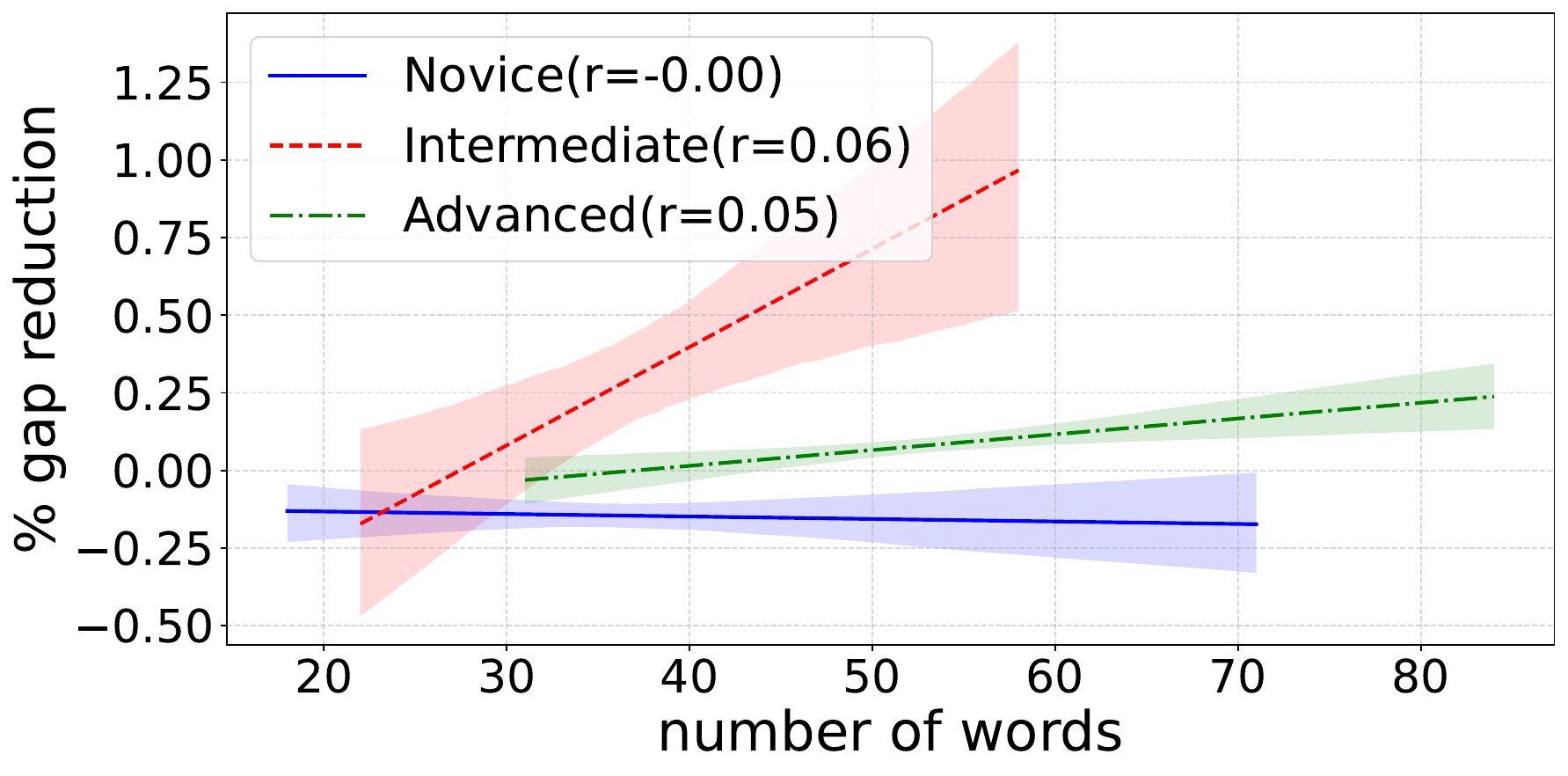}
        \label{fig:Effect_of_prompt_length}
    }
    \caption{Profiling performance by persona archetype and prompt length}
    \label{fig:Profiling_performance_by_persona_archetype_and_prompt_length}
\end{figure}

Although the MAE \emph{ultimately} converges within the range [0.3, 0.65] (Sec.~\ref{subsec:Performance_across_Domains_and_Subdomains}), Fig.~\ref{fig:MAE_across_persona_archetypes} illustrates noticeable variation during the \emph{initial} iterations, with intermediate users differing from advanced and novice users:
\begin{enumerate}[nosep, leftmargin=*]
    \item In the first iteration, the gap between the actual and predicted profiles decreases by 65\% for novice users (markedly improves with just one prompt!) and 55\% for advanced users, but increases by 75\% for intermediate users.
    \item In subsequent iterations, the MAE for novice and advanced users fluctuates and then declines, whereas for intermediate users, it stabilizes at 0.6.
\end{enumerate}

This aligns with expectations: highly and barely proficient users tend to use distinctly professional or unprofessional terms, easing profile inference. 
As noted in Sec.~\ref{subsec:Performance_across_Domains_and_Subdomains}, the impact of new (informative) prompts diminishes as $\alpha_i$ decreases.

\subsection{Performance as a Function of the Prompt Length}\label{subsec:Performance_as_a_Function_of_the_Prompt_Length}

Intuitively, longer prompts provide more cues for user profiling.  
To test this empirically, we examined how prompt length (word count) affects the reduction (\%) in the gap between actual and predicted profile scores for any assigned subdomain.  
To isolate effects, we set $\alpha_i=1$ (ignoring history, full weight on new prompts) and $|W|=0$ (no context window, considering only the prompt).  
As shown in Fig.~\ref{fig:Effect_of_prompt_length}, no statistically significant correlation was found.

\subsection{Ablation Study}\label{subsec:Ablation_Study}

To benchmark \names in the absence of closely related methods or labeled public datasets and to assess the contribution of each component, we conducted an ablation study using four variations of \names$^*$:
\begin{enumerate}[nosep, leftmargin=*]
    \item \emph{"As is"}: Default configuration with optimized hyperparameters.
    \item \emph{"$\alpha_i=0.5$"}: Fixed 50\% weight for new prompts (no decay).
    \item \emph{"$\alpha_i=1$"}: Full weight for new prompts (instantaneous learning).
    \item \emph{"Concurrent scoring"}: All assigned subdomains scored simultaneously.
\end{enumerate}

\begin{figure}[!h]
    \centering
    \subfloat[MAE analysis in the ablation study]{
        \includegraphics[width=0.47\textwidth, trim={0 7 0 8},clip]{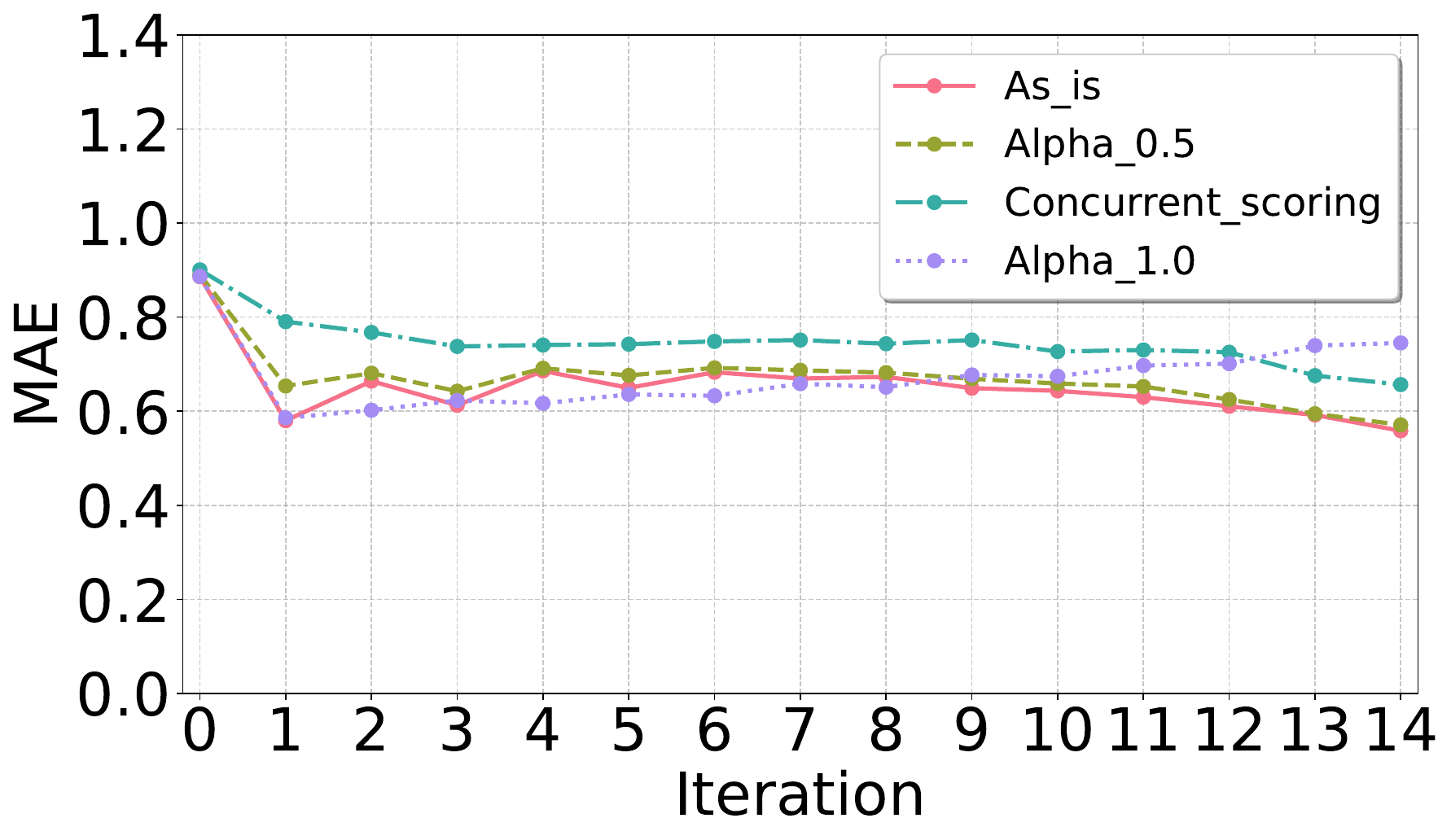}
        \label{fig:MAE_analysis_in_the_ablation_study}
    }
    \hfill
    \subfloat[MAE analysis with human users]{
        \includegraphics[width=0.47\textwidth]{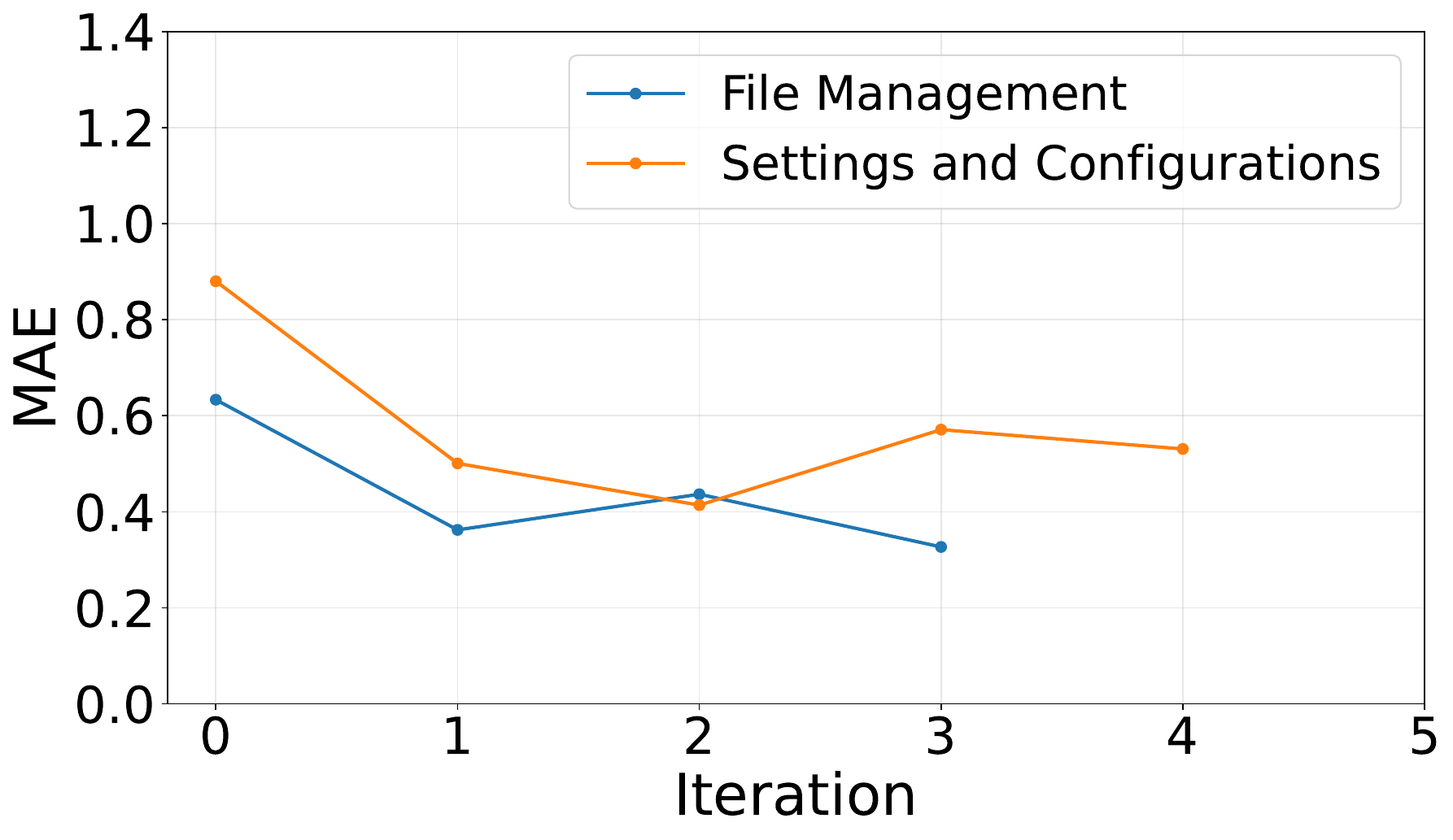}
        \label{fig:MAE_analysis_with_human_users}
    }
    \caption{Profiling performance in the ablation study and human users experiment}
    \label{fig:Profiling_performance_in_the_ablation_study_and_human_user_experiment}
\end{figure}

As shown in Fig.~\ref{fig:MAE_analysis_in_the_ablation_study}, "Concurrent scoring" performs the worst, suggesting that when a prompt is assigned to multiple subdomains, scoring should be done separately for each.  
The other variations show mixed results: some perform better in early iterations, while others improve later.  
Specifically, "$\alpha_i=1$" initially reduces MAE similarly to "As is" but then increases monotonically above all other variations.  
In contrast, "$\alpha_i=0.5$" starts off worse than "As is" but gradually converges.  
Overall, the "As is" configuration, with a gradual decrease of $\alpha_i$ and separate subdomain scoring, yields the best performance for our use case.

\subsection{Testing \names$^*$ with Human Users}\label{subsec:Testing_names_with_Human_Users}

The above experiments used synthetic users and conversations, following most prior user profiling studies~\cite{de2024use,tamoyan2024llm}.  
To complement these, we conducted an experiment with human users, who first completed our ITSec proficiency questionnaire before troubleshooting an ITSec-related scenario using chatbot assistance alone.  
In this scenario, participants faced repeated "Access denied" errors across multiple programs without intentional changes.  
Fig.~\ref{fig:MAE_analysis_with_human_users} shows that \names$^*$ assigned their prompts to either "File management" or "Settings and configurations," with prompt sequence lengths aligning with synthetic conversation distributions (Table~\ref{tab:sequence_length}).  
The resulting MAE patterns closely resemble those of synthetic users, particularly in the "Operating systems" domain (Fig.~\ref{fig:five_plus_one_MAE_domains_and_subdomains}) and the advanced user archetype (Fig.~\ref{fig:MAE_across_persona_archetypes}), showing a sharp initial improvement followed by minor fluctuations within a similar MAE range.

This experiment was limited in scale (five participants, one scenario) and involved an unrepresentative sample (graduate engineering students).  
Nevertheless, it offers reassurance regarding (1) the human-like conversational quality of synthetic users and (2) the applicability of our methods to real users.

\subsection{Qualitative Examination of User-Chatbot Conversations}\label{subsec:Qualitative_Examination_of_User_Chatbot_Conversations}

The first prompt submitted by one of the human users (Sec.~\ref{subsec:Testing_names_with_Human_Users}) was:  
\textit{"Hi there, lately I have been running into an issue in which some programs on my pc can't access their files and are getting "Access Denied" errors.
I didn't change anything.
I've check the permissions of the files and folders, but everything seems fine.
What else could it be?"}
As in other subdomains, the default $p_0^{OS/File-management}$ is 3, while this user's ground truth proficiency (from the ITSec questionnaire) is 2.66.  
The user provided a clear description, mentioning relevant technical terms (programs, permissions, files) and demonstrating more than basic proficiency by checking permissions.  
However, the user lacked deeper expertise for advanced diagnosis beyond standard UI-based troubleshooting.  
Accordingly, \names$^*$ assigned $p_1^{OS/File-management, temp}$ =2.75, and after applying Eq.~\ref{eq:score_update}, updated $p_1^{OS/File-management}$ to 2.8.  
A qualitative review of two other conversations is available in a video demonstration we produced\footnote{https://tinyurl.com/2a76bbvb}.

\subsection{Overhead}\label{subsec:Overhead}

\begin{table}[t]
\centering
\caption{Distribution of the overhead associated with our experiments}
\resizebox{\columnwidth}{!}{%
\begin{tabular}{l|l|c|c|c|c|c}
\toprule
 & & \textbf{User generation} & \textbf{Chatbot conversation} & \textbf{Conversation QA} & \multicolumn{2}{c}{\textbf{Profiling}} \\
\cmidrule(lr){6-7}
 & & & & & \textbf{Subdomain assignment} & \textbf{Subdomain Scoring} \\
\midrule
Seconds & & 1.35$\pm$ 5.23& 6.59 $\pm$ 8.23 & 8.09 $\pm$ 12.88 & 1.1 $\pm$ 11.66 & 2.4 $\pm$ 15.73 \\
\midrule
\multirow{2}{*}{Tokens} & Input & 938.77 $\pm$ 108.26 & 3,372.71$\pm$ 669.13 & 4,040.92$\pm$193.99& 504.74 $\pm$ 118.25 & 1,363 $\pm$ 248.29 \\
\cmidrule(lr){2-7}
 & Output & 285.71 $\pm$ 35.26 & 699.13$\pm$511.12 &468.64$\pm$16.94 & 25.28 $\pm$ 8.23 & 67.13 $\pm$ 2.92 \\
\bottomrule
\end{tabular}%
}
\label{tab:overhead}
\end{table}

As detailed in Sec.~\ref{subsec:ProfiLLM_Implementation}, our experiments were conducted on Azure, providing access to multiple state-of-the-art pretrained LLMs.  
Using these LLMs incurs costs, and \names's LLM-based data preprocessing and analysis introduce time overhead.  
Table~\ref{tab:overhead} presents the latency (seconds) and monetary cost (token count) for data generation, refinement, and profiling.  
Except for user generation (once per user) and conversation QA (once per generated conversation), the values ($\text{Mean} \pm \text{St.Dev.}$) represent a single user prompt.  
Notably, the most resource-intensive component -- both in time and tokens -- was the conversation quality assurance (Sec.~\ref{subsec:Conversation_Data_Collection_and_Refinement}) performed by Claude.  
All other tasks used GPT-4o, which efficiently generated users from numeric profile vectors (Sec.~\ref{subsec:Synthetic_User_Generation}).  
As expected, chatbot interaction generation with synthetic users required more resources.  
For the LLM-based profiling components, Table~\ref{tab:overhead} demonstrates practical feasibility, with low latency and affordable operation costs.

\section{Related Work}\label{sec:Related_Work}

\subsection{User Profiling Dimensions and Taxonomies}\label{subsec:User_Profiling_Dimensions_and_Taxonomies}
Previous research has explored various user profile dimensions.  
\emph{Personality} is a well-studied aspect, particularly through the Big Five Factors (BFF) framework~\cite{digman1990personality}, which includes openness, conscientiousness, extroversion, agreeableness, and neuroticism.  
Inferring these traits enables chatbots to enhance personalization~\cite{meduri10adaptive}.  
User profiling has also been applied in e-commerce~\cite{liu2024once}, education~\cite{Koyuturk2023Developing} and healthcare~\cite{Ghanadian2023llChatGPT} for user preference modeling, activity adaptation and mental health assessment, respectively.  
Closer to the ITSec domain addressed in our study, Bitton et al.~\cite{BITTON2018266} proposed a taxonomy and a method to assess mobile security awareness.  
However, we found no prior work on inferring technical proficiency, particularly in ITSec, solely from conversational data.

\subsection{User Profiling Inputs and Methods}\label{subsec:User_Profiling_Inputs_and_Methods}
Various methods for user profile inference leverage diverse data sources and techniques from machine learning~\cite{g2021survey}, deep learning~\cite{qian2021learning}, and NLP~\cite{flores2022user}.  
Recently, LLMs have enhanced user profiling by improving the interpretation of user-generated content~\cite{tan2023user}.  
Several studies have explored personality detection from text using LLMs, including transformer-based models~\cite{jain2022personality}, ChatGPT prompting~\cite{de2024use}, and fine-tuned pretrained models.  
Unlike personality profiling~\cite{wen2023desprompt}, typically evaluated on static text datasets, or preference modeling~\cite{liu2024once}, based on past interactions and demographics, \name is tailored for conversational settings.  
It infers user profiles implicitly and dynamically from sequences of user prompts and chatbot responses, leveraging a domain-adapted taxonomy.

\subsection{User Profiling in Conversational Settings}\label{subsec:User_Profiling_in_Conversational_Settings}

Interaction-based personalization has gained significant attention, enabling chatbots to adapt responses to a user's communication style and preferences~\cite{liu2017content}.  
With LLM advancements, profiling has shifted to dynamic and implicit methods, allowing real-time inference from interactions (as in \name).  
For example, DHAP~\cite{ma2021one} learns implicit user profiles from past conversations to personalize chatbot responses.  
Unlike DHAP, which focuses on language style and relies on historical logs, \name specializes in domain-specific expertise profiling and continuously updates profiles during live interactions.  
Gandoul et al.~\cite{gandoultowards} explore chatbot-based inference of students' learning styles, but their work remains high-level, lacking experiments, code, or data.  
Apollonion~\cite{chen2024apollonion}, like \name, dynamically builds structured user profiles but incorporates pre-existing user data, whereas \name relies solely on conversational inputs.  
Moreover, \name is reproducible, with publicly shared code and datasets.  

Overall, unlike prior research focused on narrow profile dimensions (e.g., personality), \name supports diverse domain-adapted expertise assessments.  
Profiling users based solely on conversational data, particularly in specialized fields like ITSec, presents greater challenges than widely studied personality inference.  
Instead of relying on static datasets, \name dynamically infers context-aware profiles, enhancing adaptability in knowledge-intensive domains.

\section{Discussion}\label{sec:Discussion}

\subsection{Impact and Generalizability}\label{subsec:Impact_and_Generalizability}

As noted in Sec.~\ref{sec:Related_Work}, user profiling -- often personality-oriented -- allows chatbots to adapt the \emph{style and tone} of responses, enhancing user satisfaction.
With \names, designed for ITSec troubleshooting, other benefits emerge, such as optimizing the \emph{choice of terms} as well as the \emph{number and complexity of steps} per solution.  
For instance, an advanced user may be directed to inspect the Task Manager for suspicious processes, while a novice receives step-by-step instructions in simpler terms.  
These benefits extend to other domains, such as Legal \& Compliance (adjusting terminology and complexity based on legal expertise) and Data Science (tailoring explanations of models, algorithms, and best practices).  
Adapting \name to a new domain primarily involves developing a relevant taxonomy and refining system prompts accordingly.

\subsection{Limitations and Future Work}\label{subsec:Limitations_and_Future_Work}

This study focused on optimizing profile inference, leaving profile-aware response adaptation as a challenge for future research.  
A key question is whether adapting chatbot responses based on inferred profiles influences user behavior and profiling accuracy.  
Another limitation is the deliberate design of \name as \emph{fully} implicit and non-disruptive.  
Relaxing this constraint could enhance performance -- incorporating challenge questions or preliminary questionnaires may improve profiling accuracy, an aspect for future evaluation.  
Extending \name to new domains and testing it with larger human user groups is a key research direction.

\section{Conclusion}\label{sec;Conclusion}

We introduced \name, an LLM-based method for real-time, non-disruptive profiling of chatbot users using conversational prompts alone.  
\name enables structured, fully implicit profiling that dynamically adapts to user interactions, enhancing adaptability in conversational AI.  
Additionally, in evaluating \name, we (1) proposed a novel taxonomy for ITSec proficiency profiling,  
(2) developed and empirically tested \names, an ITSec-adapted variant of \name, and  
(3) developed an LLM-driven framework for generating synthetic users that engage in profile-consistent chatbot interactions while ensuring quality.  
Experiments demonstrated favorable performance trends for \names, particularly a rapid and steep MAE reduction, enabling early profile-aware response adaptation.  
Unlike most existing methods, \name allows refining response terminology and complexity rather than style and tone.

\bibliographystyle{plain}
\bibliography{bib}

\end{document}